\documentclass{article}

\usepackage{amsmath,amssymb}
\usepackage{microtype}
\usepackage{graphicx}
\usepackage{booktabs} 
\usepackage{subcaption}
\usepackage[normalem]{ulem}
\usepackage{multicol}
\usepackage{tabularx, makecell, multirow} 

\usepackage{hyperref}

\usepackage[accepted]{icml2021}

\icmltitlerunning{How Do Adam and Training Strategies Help BNNs Optimization?}

\begin{document}

\twocolumn[
\icmltitle{How Do Adam and Training Strategies Help BNNs Optimization?}
~
\icmlsetsymbol{equal}{*}

\begin{icmlauthorlist}
\icmlauthor{Zechun Liu}{equal,hkust,cmu}
\icmlauthor{Zhiqiang Shen}{equal,cmu}
\icmlauthor{Shichao Li}{hkust}
\icmlauthor{Koen Helwegen}{plumerai}
\icmlauthor{Dong Huang}{cmu}
\icmlauthor{Kwang-Ting Cheng}{hkust}
\end{icmlauthorlist}

\icmlaffiliation{hkust}{Hong Kong University of Science and Technology}
\icmlaffiliation{cmu}{Carnegie Mellon University}
\icmlaffiliation{plumerai}{Plumerai}
\icmlcorrespondingauthor{Zhiqiang Shen}{zhiqians@andrew.cmu.edu}
\icmlcorrespondingauthor{Zechun Liu}{zechun.liu@connect.ust.hk}

\vskip 0.3in
]

\printAffiliationsAndNotice{\icmlEqualContribution} 

\begin{abstract}
The best performing Binary Neural Networks (BNNs) are usually attained using Adam optimization and its multi-step training variants~\cite{rastegari2016xnor,liu2020reactnet}. However, to the best of our knowledge, few studies explore the fundamental reasons why Adam is superior to other optimizers like SGD for BNN optimization or provide analytical explanations that support specific training strategies.
To address this, in this paper we first investigate the trajectories of gradients and weights in BNNs during the training process. We show the regularization effect of second-order momentum in Adam is crucial to revitalize the weights that are dead due to the activation saturation in BNNs. We find that Adam, through its adaptive learning rate strategy, is better equipped to handle the rugged loss surface of BNNs and reaches a better optimum with higher generalization ability. 
Furthermore, we inspect the intriguing role of the real-valued weights in binary networks, and reveal the effect of weight decay on the stability and sluggishness of BNN optimization. Through extensive experiments and analysis, we derive a simple training scheme, building on existing Adam-based optimization, which achieves 70.5\% top-1 accuracy on the ImageNet dataset using the same architecture as the state-of-the-art ReActNet~\cite{liu2020reactnet} while achieving 1.1\% higher accuracy. Code and models are available at \url{https://github.com/liuzechun/AdamBNN}.

\end{abstract}

\section{Introduction}
\label{sec:intro}

Binary Neural Networks (BNNs) have gained increasing attention in recent years due to the high compression ratio~\cite{rastegari2016xnor} and the potential of being accelerated with logic computation on hardware~\cite{zhang2019dabnn}. Their applications range from supervised learning, e.g., classification~\cite{courbariaux2016binarized}, segmentation~\cite{zhuang2019structured}, pose estimation~\cite{bulat2019improved} to the self-supervised learning~\cite{shen2021s2}. 

\begin{figure}[t]
	\centering
	    \includegraphics[width=0.48\textwidth]{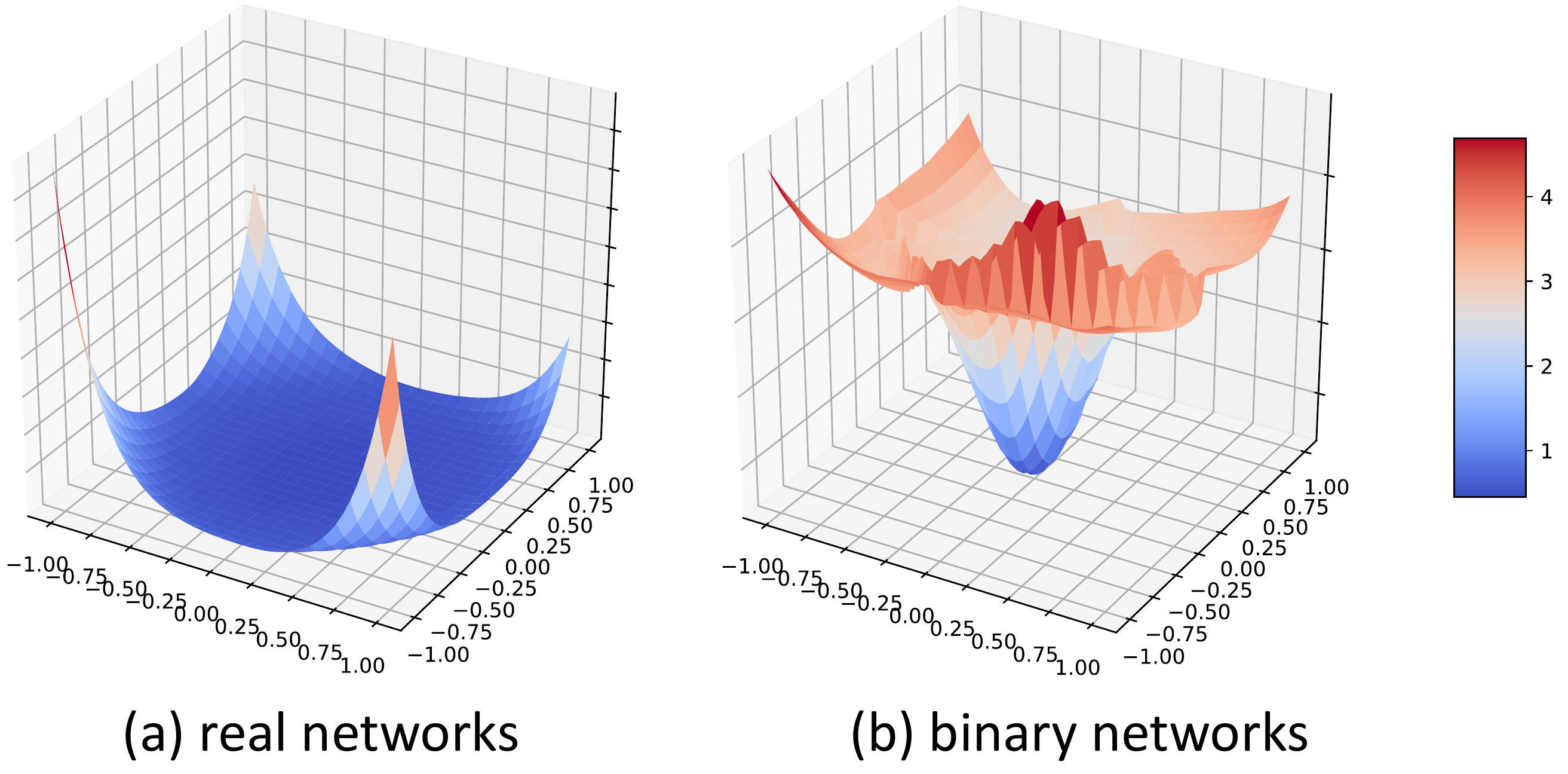}
	\vspace{-2.2em}
	\caption{The actual optimization landscape from real-valued and binary networks with the same architecture (ResNet-18). We follow the method in~\cite{li2018visualizing} to plot the landscape.} 
	\label{fig:landscape}
	\vspace{-0.5em}
\end{figure}

Despite the high compression ratio of BNNs, the discrete nature of the binary weights and activations poses a challenge for its optimization. 
It is widely known that conventional deep neural networks rely heavily on the ability to find good optima in a highly non-convex optimizing space. Different from real-valued neural networks, binary neural networks restrict the weights and activations to discrete values ({-1, +1}), which naturally, will limit the representational capacity of the model and further result in disparate optimization landscapes compared to real-valued ones. 
As illustrated in Figure~\ref{fig:landscape}, BNNs are more chaotic and difficult for optimization with numerous local minima compared to real-valued networks. These properties differentiate BNNs from real-valued networks and impact the optimal optimizer and training strategy design.

Since Courbariaux et al.~\cite{courbariaux2016binarized} adopted Adam as the optimizer for BNNs, multiple researchers independently observed that better performance could be attained by Adam optimization for BNNs~\cite{bethge2020meliusnet,liu2020reactnet,martinez2020training}. However, few of these works have analyzed the reasons behind Adam's superior performance over other methods, especially the commonly used stochastic gradient descent (SGD)~\cite{robbins1951stochastic} with first momentum.

\begin{figure}[t]
	\centering
	\begin{subfigure}[b]{0.242\textwidth}
	\centering
	    \includegraphics[width=\textwidth]{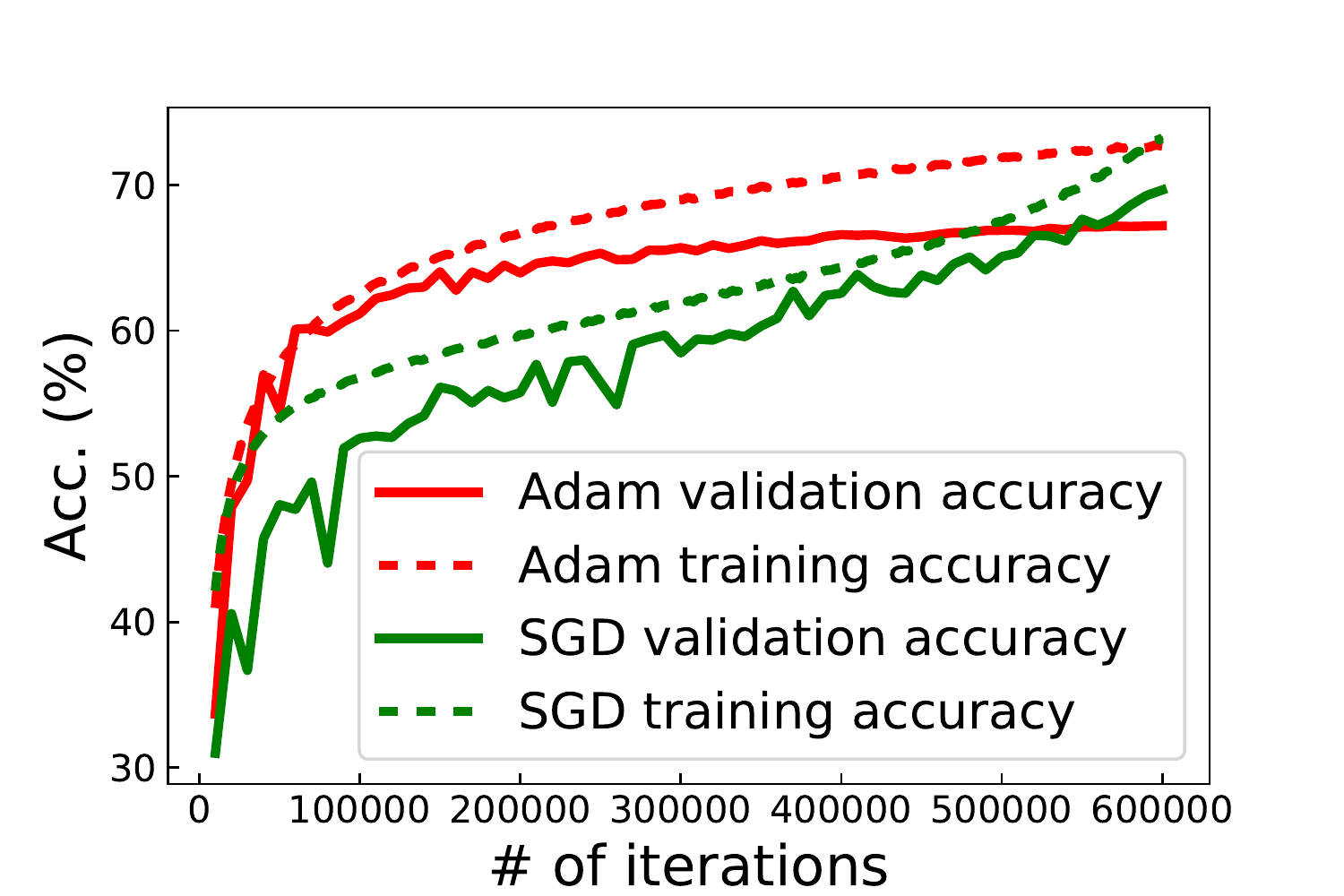}
	    \vspace{-1.5em}
	    \caption{real networks}
	\end{subfigure}%
	\begin{subfigure}[b]{0.242\textwidth}
	\centering
	    \includegraphics[width=\textwidth]{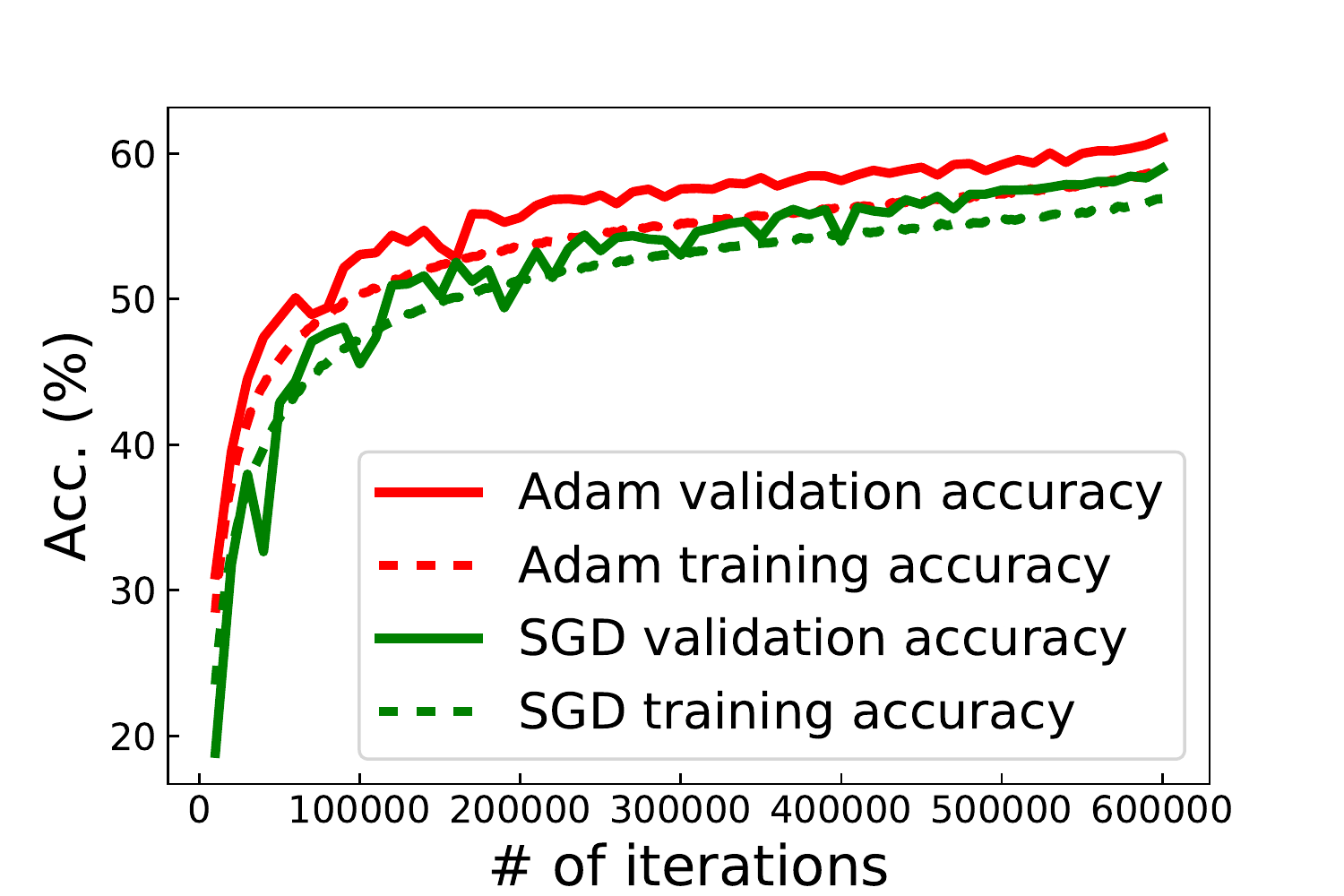}
	    \vspace{-1.5em}
	    \caption{binary networks}
	\end{subfigure}%
	\vspace{-1em}
	\caption{The top-1 accuracy curves of the real-valued and binary network (ResNet-18 based) trained on ImageNet. On the real-valued network, SGD achieves higher accuracy with better generalization ability in the final few iterations. Binarization has a strong regulating effect, resulting in the validation accuracy being higher than the training accuracy. Adam outperforms SGD under this circumstance.} 
	\label{fig:accuracy_curve}
\end{figure}

Recent theoretical work from Wilson et al.~\cite{wilson2017marginal} empirically shows that adaptive learning rate methods like Adam reach fewer optimal minima than SGD with momentum, meaning that minima found by SGD generalize better than those found by Adam.
This matches experience with real-valued neural networks where state-of-the-art results for many tasks in Computer Vision~\cite{tan2019efficientnet} and Machine Translation~\cite{wu2016google} are still obtained by plain SGD with momentum. It seems counter-intuitive considering that Adam comes with better convergence guarantee and should deliver better performance over SGD. We observe the real-valued networks are fairly powerful to ``overfit'' on the training data as shown in Figure~\ref{fig:accuracy_curve} (a), but as we will demonstrate later, this may be not true for BNNs. We observe that BNNs are usually under-fitting on the training set due to the limited model capacity (the performance on validation is higher than that on training set), even if we train BNNs thoroughly with a longer budget. From Figure~\ref{fig:accuracy_curve} (b), it is evident that the validation accuracy of SGD on binary networks fluctuates more compared to Adam, which indicates that on binary training, SGD easily gets stuck in the rugged surface of the discrete weight optimization space and fails to find generalizable local optima. 

Based on these observations, in this paper, we investigate the fundamental reasons why Adam is more effective for BNNs than SGD. During BNN training, a proportion of gradients tends to be zero due to the activation saturation effect. If using SGD as the optimizer, the updating step of the individual weight aligns with the corresponding gradient in the magnitude, making it hard to flip those ``dead'' weights out of a bad initialization or local minima. Intuitively, revitalizing the ``dead'' weights with appropriate gradients can significantly improve the accuracy of BNNs, which is further supported by our visualization results and the final accuracy. According to our experiments, the normalization effect from the second momentum of Adam rescales the updating value element-wisely based on the historical gradients, effectively resolving the ``dead'' weights issue. 

Besides comparing Adam to SGD, we further explore how training strategies affect BNN optimization. Previous works proposed different training strategies: Yang et al.~\cite{yang2019synetgy} proposed to progressively quantize the weights from 16 bits to 1 bit. Zhuang et al.~\cite{zhuang2018towards} proposed binarizing weights first and binarizing activations in the second step. More recently Martinez et al.~\cite{martinez2020training} proposed a two-step strategy to binarize activations first and then binarize weights. These works involve complex training strategy designs but seldom explain the reason behind those designs. Instead of proposing a new training strategy, in the second part of our work, we explain the mechanisms behind BNN training strategies, from an important but overlooked angle -- weight decay. We quantify the effects of weight decay on the BNN optimization’s stability and initialization dependency with two metrics, FF ratio and C2I ratio, respectively. Guided by these metrics, we identify a better weight decay scheme that promotes the accuracy of the state-of-the-art ReActNet from 69.4\% to 70.5\%, surpassing all previously published studies on BNNs. 

Unlike previous studies that focus on designing network architectures for BNNs, we focus on the investigation of optimizers and training strategy, which we think is valuable for maximizing the potential in a given structure for better performance.
All of our experiments are conducted on the full ImageNet\footnote{Several previous works conduct experiments on small datasets like MNIST and CIFAR-10/100, and sometimes draw conclusions that are inconsistent with experiments on large-scale/real-world.\vspace{-1em}}, which is more reliable. We believe our exploratory experiments will be beneficial for the research on BNNs optimization and may inspire more interesting ideas along this direction.

\noindent{\textbf{Contributions.}}
In summary, we address the following issues and our contributions are as follows:
\vspace{-0.06in}

$\bullet$ We provide thorough and fair comparisons among different optimizers for BNN optimization, especially between Adam and SGD, on the large-scale ImageNet dataset. We further design several metrics to analyze the patterns beneath the binary behavior and present a simple visualization method based on the alteration of gradients and weights inside training.
\vspace{-0.06in}

$\bullet$ We explain the difficulties that arise from a non-adaptive learning rate strategy by visualizing these trajectories and show that optimization lies in extremely rugged surface space. We conclude that gradient normalization is crucial for BNN optimization.
\vspace{-0.06in}

$\bullet$ We further examine the existing practice in BNN optimization strategy design and provide in-depth analysis on the weight decay effect. Based on these analyses, we propose practical suggestions for optimizing BNNs. These techniques help us to train a model with 1.1\% higher accuracy than the previous state-of-the-art results. 

\section{Related Work}
Research on binary neural network optimization can be mainly divided into several aspects:

\vspace{-0.2cm}
\textbf{Structure Adjustment}
Previous attempts to improve BNNs are mainly paid on network structure design, including adding real-valued shortcuts~\cite{liu2018bi,liu2018bi_journal,liu2020reactnet}, or real-valued attention blocks~\cite{martinez2020training}, expanding the channel width~\cite{mishra2017wrpn, zhuang2019structured}, ensemble more binary networks~\cite{zhu2019binary} or use a circulant convolution~\cite{liu2019circulant}. These works provide advanced structures that bring breakthroughs in accuracy. In this work, we are motivated to disambiguate the binary optimization process, which is orthogonal to the structural design. 

\vspace{-0.2cm}
\textbf{Gradient Error Reduction and Loss function Design}
Some studies pay attention to reduce the gradient error of the BNNs, for example, XNOR-Net~\cite{rastegari2016xnor} uses a real-valued scaling factor multiplying with the binary weights and activations, and ABC-Net~\cite{lin2017abcnet} adopts more weight bases. IR-Net~\cite{qin2020forward} propose Libra-PB to simultaneously minimize both quantization error and information loss. A few works adjust the loss functions. Hou et al. proposed loss-aware binarization~\cite{hou2016loss} using the proximal Newton algorithm with the diagonal Hessian approximation to directly minimize the loss w.r.t. binary weights. Ding et al. proposed activation regularization loss to improve BNN training~\cite{ding2019regularizing}. These studies also aim to resolve the discreteness-brought optimization challenge in binary neural networks. Instead, we scrutinize another important yet less investigated angle, the optimizer and optimization strategy reasoning.

\vspace{-0.2cm}
\textbf{Optimizer Choice and Design}
Recently, many binary neural network choose Adam over SGD, including BNN~\cite{courbariaux2016binarized}, XNOR-Net~\cite{rastegari2016xnor}, Real-to-Binary Network~\cite{martinez2020training}, Structured BNN~\cite{zhuang2019structured}, ReActNet~\cite{liu2020reactnet}, etc. Helwegen et al. proposed a new binary optimizer design based on Adam~\cite{helwegen2019latent}. Empirical studies of binary neural network optimization~\cite{alizadeh2018empirical, tang2017train} also explicitly mention that Adam is superior to SGD and other optimization methods. However, the reason why Adam is suitable for binary network optimization is still poorly understood. In this study, we investigate the behavior of Adam, attempting to bring attention to the binary optimizer understanding and improving the binary network performance within a given network structure, which we hope is valuable for the community.

\vspace{-0.2cm}
\textbf{Training Strategy}
Multiple works proposed different multi-step training strategies to enhance the performance of BNNs. Zhuang et al.~\cite{zhuang2018towards} proposed to first quantize the weights then quantize both weights and activations. Following~\cite{zhuang2018towards}, Yang et al.~\cite{yang2019synetgy} proposed to progressively quantize weights and activations from higher bit-width to lower bit-width. Recent studies~\cite{martinez2020training, liu2020reactnet} proposed to binarize activation first, then in the second stage, further binarize the weights. Those previous work each proposed their own training techniques, but seldom generalized techniques into the reasons behind, which also brings confusion to followers in determining which technique they can use in their circumstance. In this work, we analyze the foundations of choosing optimization strategies beyond providing a possible solution, in hope of inspiring more interesting solutions in this area. 

\section{Methodology}

This section begins by introducing several observations from real-valued networks and binary neural networks (BNNs) training. We observe the generalization ability of Adam is better than SGD on BNNs, as shown in Figure~\ref{fig:accuracy_curve}. This phenomenon motivates us to ask why SGD works better for a real-valued classification network, but loses its superiority in binary neural network optimization. Start by this, we visualize the activation saturation phenomenon during optimizing an actual binary neural network and deliberate its effects on gradient magnitude in Section~\ref{sec:saturation}. Then we observe that activation saturation will cause the unfair training problem on channel weights as described in Section~\ref{sec:fairness}. Further, for clear explanation we construct an imaginary two-dimensional loss landscape containing sign functions to mimic a simplified optimization process of BNNs with activation binarization in Section~\ref{sec:2D_loss}, and we analysis how Adam can help conquer the zero-gradient local minima. Moreover, we point out that the real-valued weights in BNNs can be regarded as the {\em confidence} score, as described in Section~\ref{sec:latent_weight}, making BNN optimization intricate. Thus, we define several metrics in Section~\ref{sec:metrics} to depict the property of BNNs and measure the goodness of a BNN training strategy. Lastly, we provide practical suggestions for optimizing BNNs.

\begin{figure}[t]
	\centering
	    \includegraphics[width=0.48\textwidth]{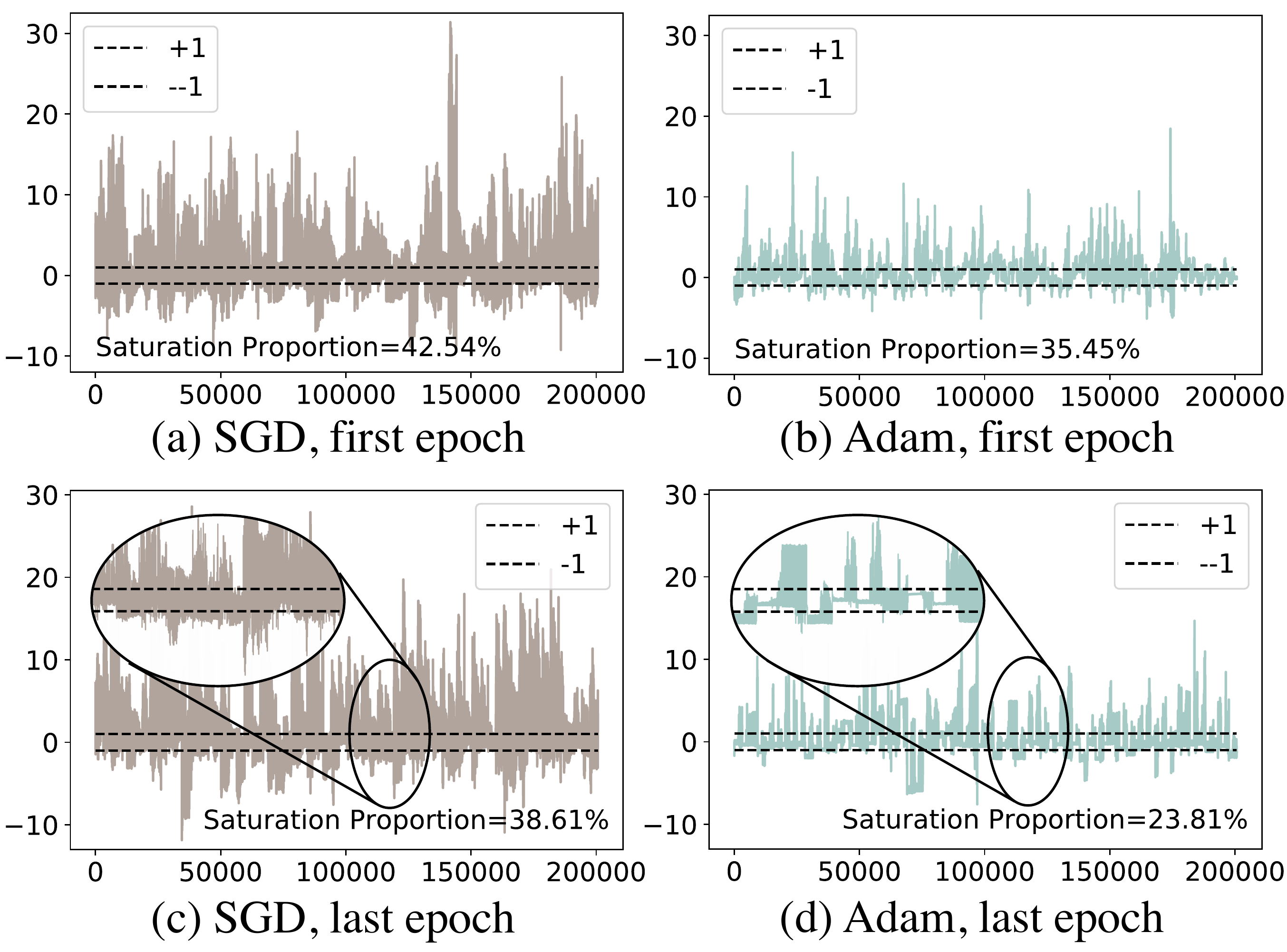}
	\vspace{-2.2em}
	\caption{Activation distributions in binary ResNet-18 structure from different optimizers on ImageNet. Dotted lines are the up (+1) and low (-1) bounds. We plot the input activation to the first binary convolution and we observe that both SGD and Adam optimized BNNs experienced activation saturation. However, Adam can alleviate activation saturation during optimization compared to SGD, as shown in the zoom-in views in (c) and (d). We further count the number of activations that are over the bounds for SGD and Adam, the percentages are 42.54\% and 35.45\% respectively after the first epoch, 38.61\% and 23.81\% after the last epoch. The activation saturation proportion from Adam optimization is significantly lower than SGD. More details please refer to Section~\ref{sec:saturation}.} 
	\label{fig:activation_saturation}
\end{figure}

\begin{figure*}[t]
	\centering
	\begin{subfigure}[b]{0.33\textwidth}
	\centering
	    \includegraphics[height=0.535\textwidth]{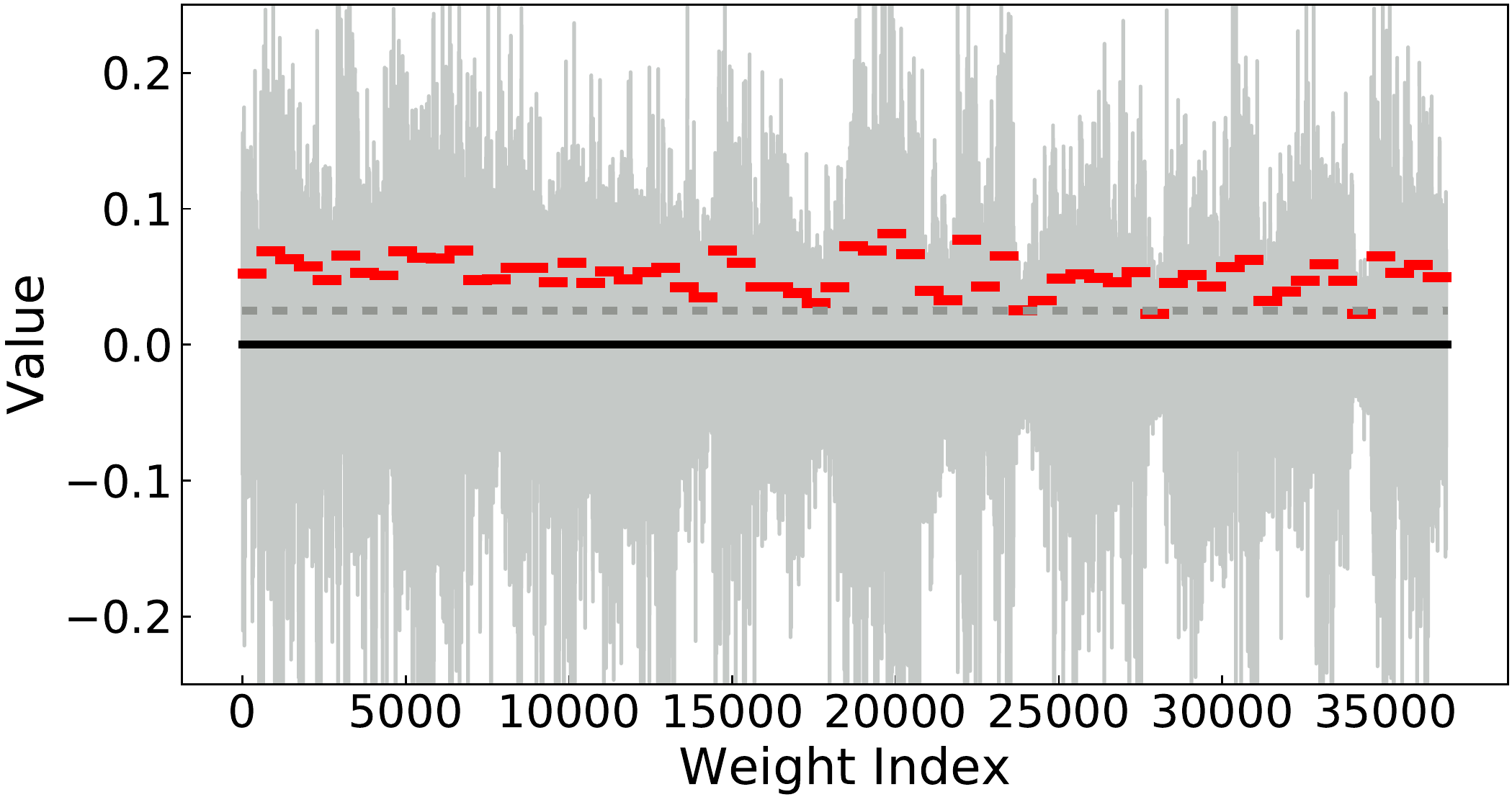}
	    \caption{Real-valued network with SGD}
	\end{subfigure}%
	\begin{subfigure}[b]{0.33\textwidth}
	\centering
	    \includegraphics[height=0.535\textwidth]{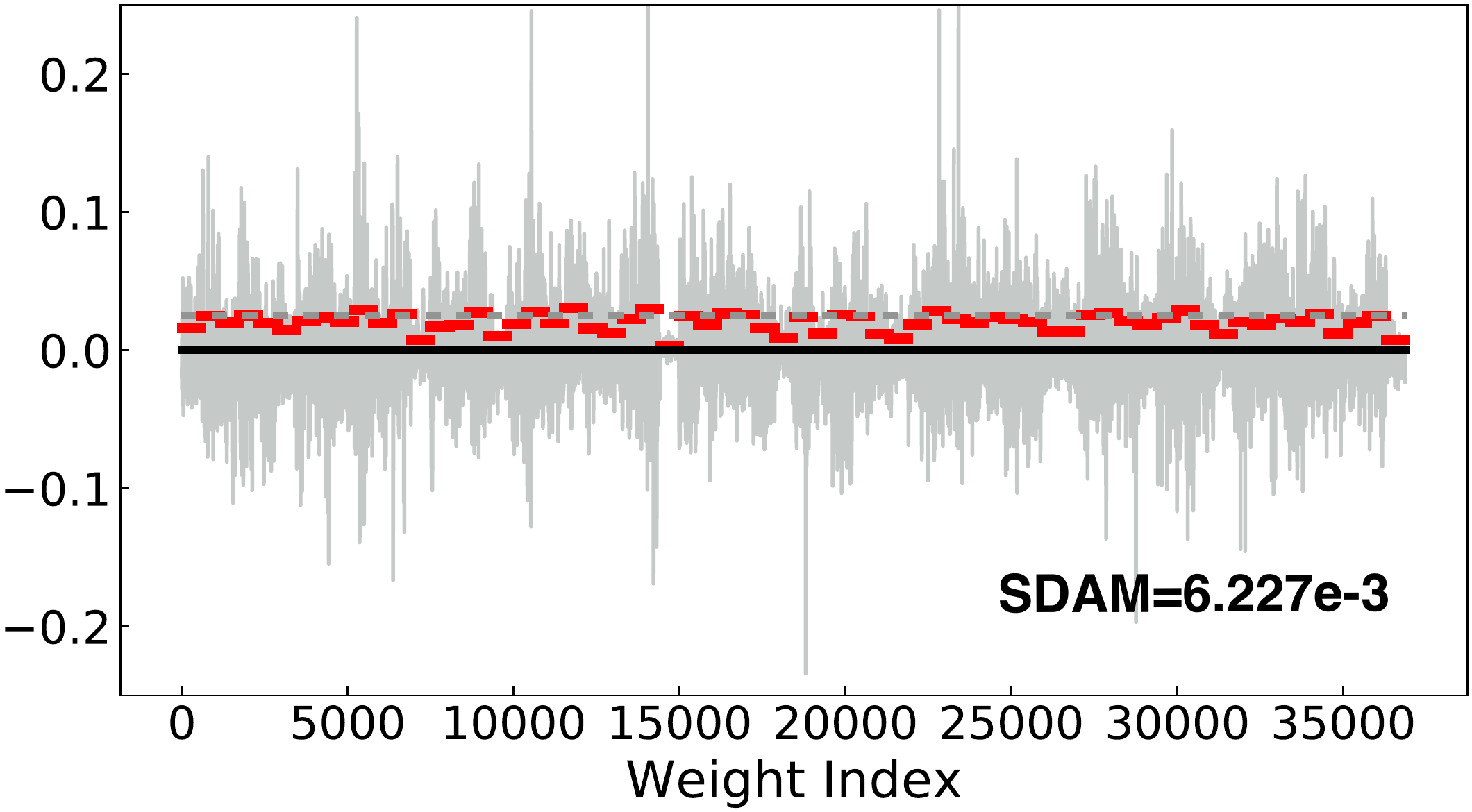}
	    \caption{Binary network with SGD}
	\end{subfigure}%
	\begin{subfigure}[b]{0.33\textwidth}
	\centering
	    \includegraphics[height=0.535\textwidth]{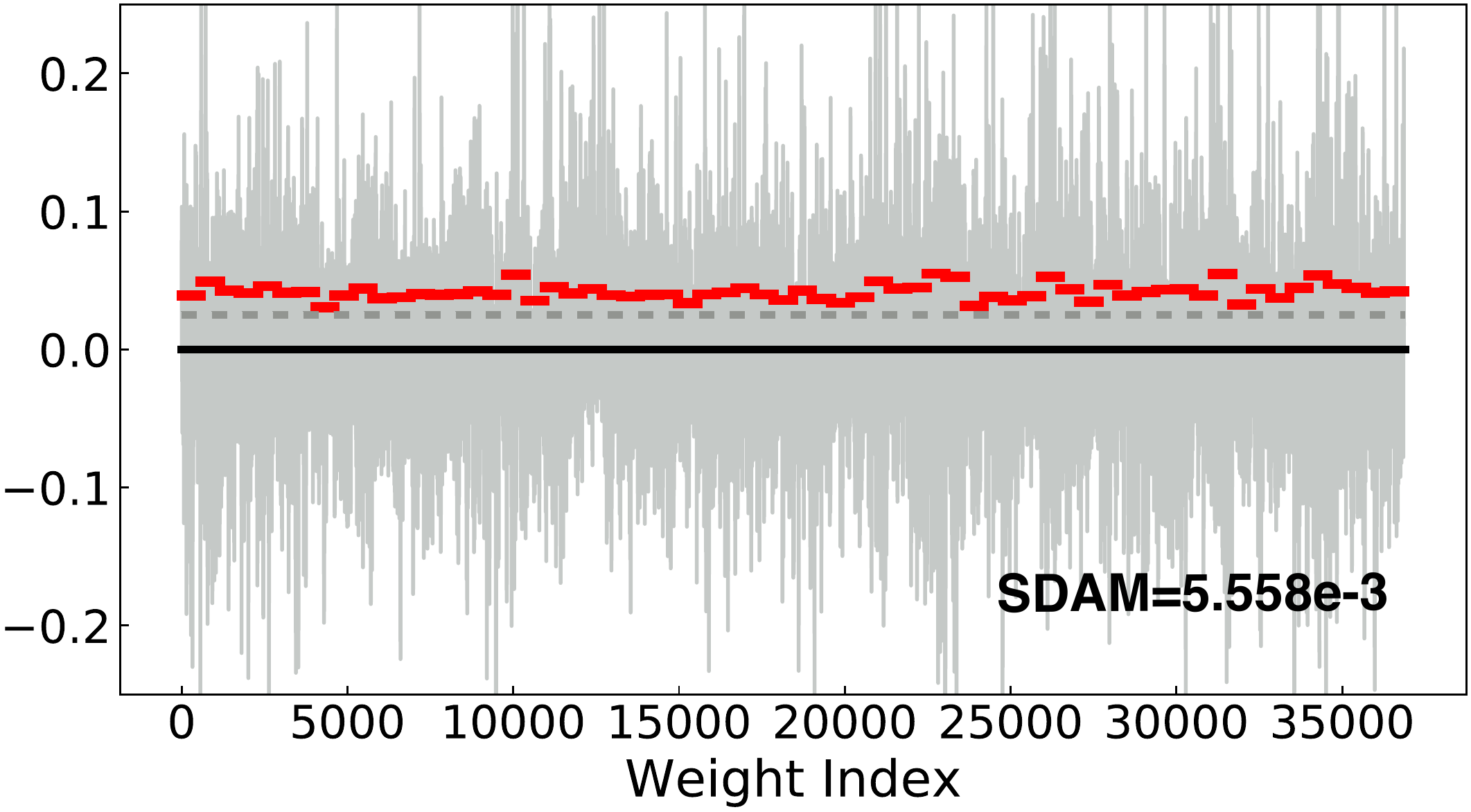}
	    \caption{Binary network with Adam}
	\end{subfigure}%
	\vspace{-1em}
	\caption{The weight value distribution in the first binary convolutional layer after training one epoch. For clarity, we use \textit{red} hyphens to mark the Channel-wise Absolute Mean (CAM) of real-valued weights in each kernel. The \textit{grey} dotted line denotes the minimum CAM value (0.0306) of weights in the Adam optimized binary network. Compared to Adam, SGD optimization leads to much lower CAM value, and higher Standard Deviation (SDAM), which indicates that the weights optimized with SGD are not as fair (well-trained) as those with Adam. More detailed analysis can be found in Section~\ref{sec:adam>sgd}.}
	\label{fig:weight_distribution}
\end{figure*}

\begin{figure*}[t]
	\centering
	\begin{subfigure}[b]{0.48\textwidth}
	\centering
	    \includegraphics[height=0.54\textwidth]{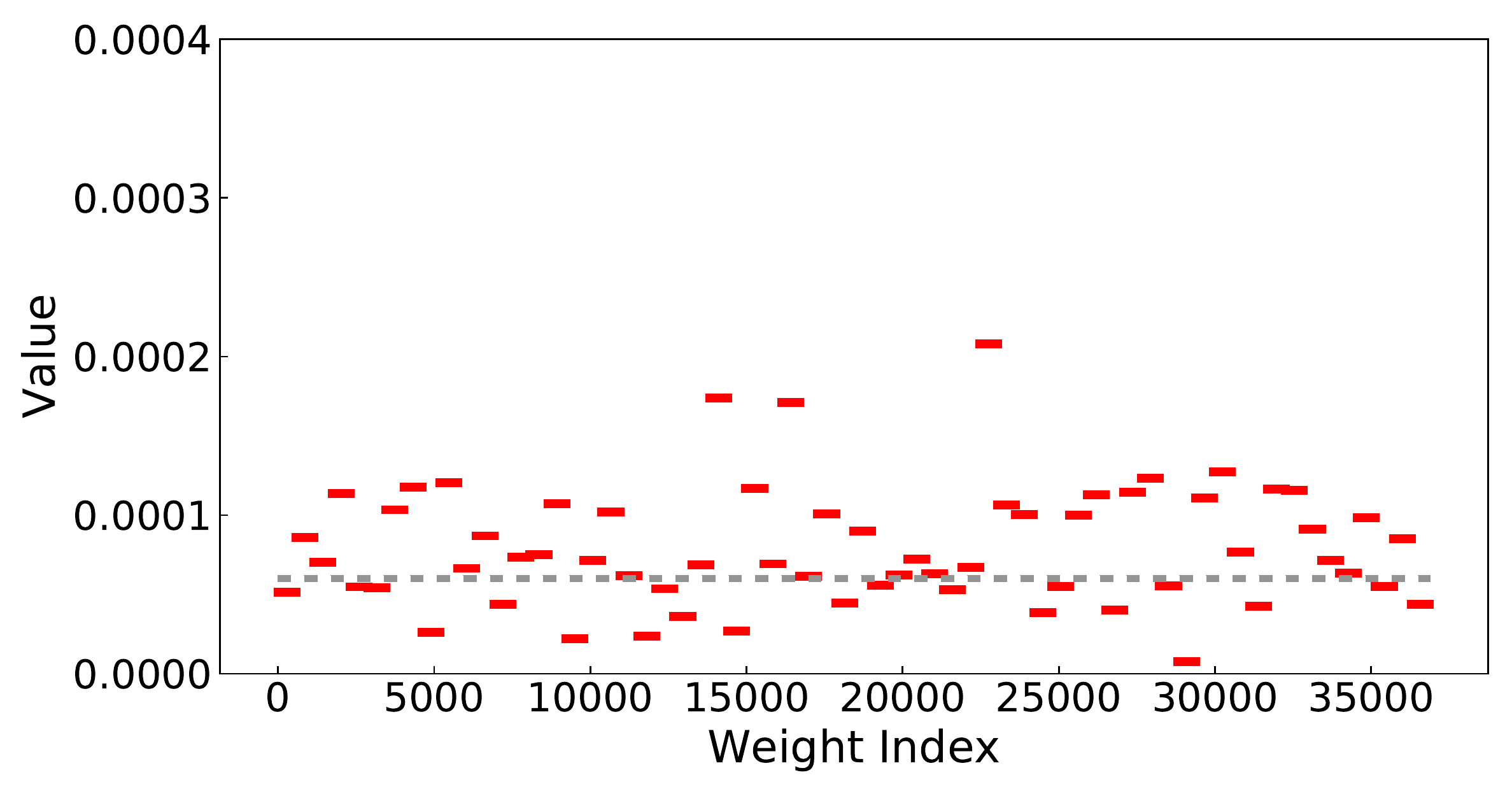}
	    \vspace{-1.8em}
	    \caption{Binary network with SGD}
	\end{subfigure}
	\begin{subfigure}[b]{0.48\textwidth}
	\centering
	    \includegraphics[height=0.54\textwidth]{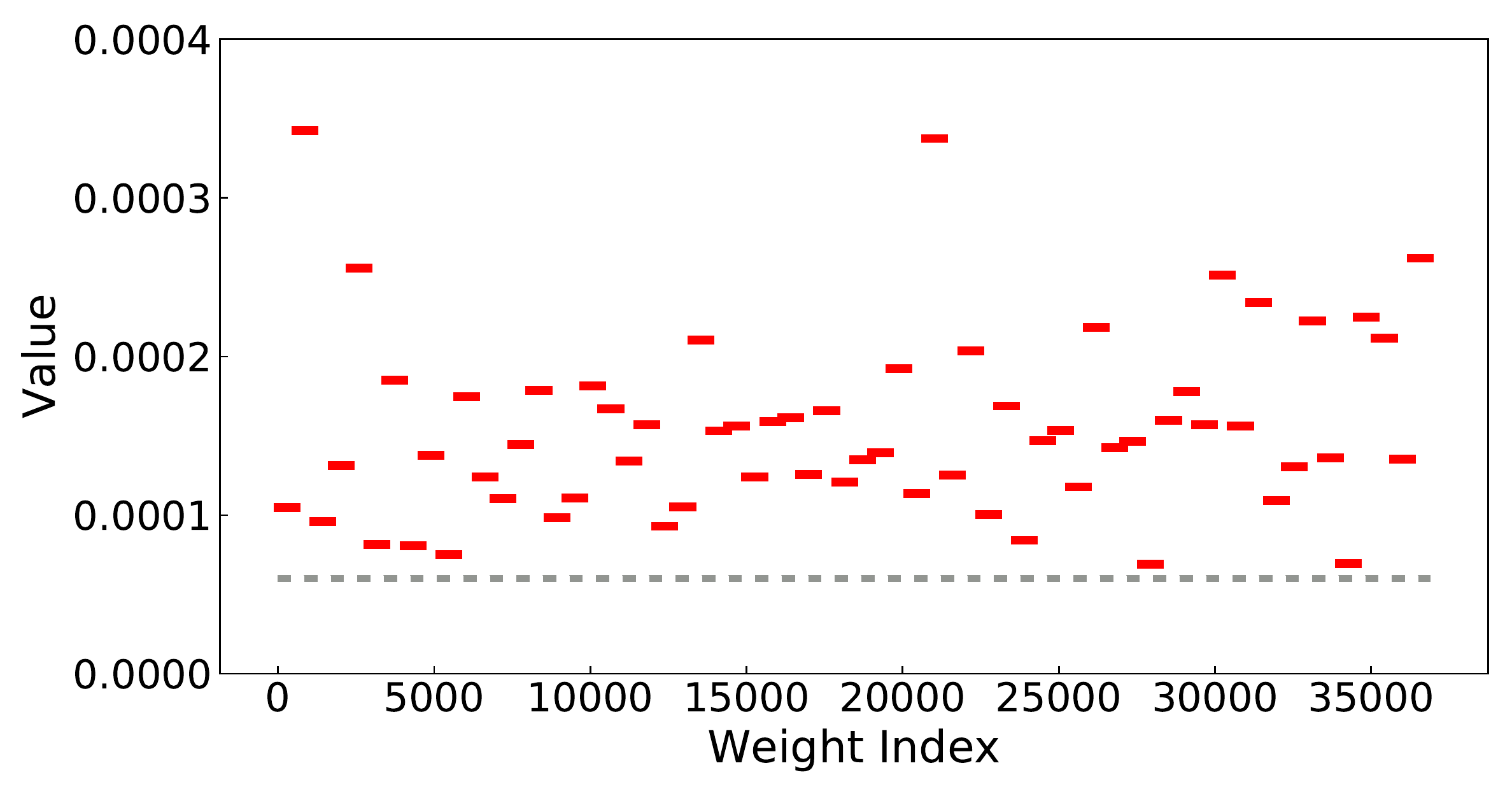}
	    \vspace{-1.8em}
	    \caption{Binary network with Adam}
	\end{subfigure}
	\vspace{-1em}
	\caption{The update value distribution of weights in the first binary convolutional layer after trained with one epoch. For clarity, we omit the original update value distribution and use \textit{red} hyphens to mark the Channel-wise Absolute Mean (CAM) of the weights' update values in each kernel. In this layer, 34.3\% of the kernels in SGD have a lower CAM than the minimum CAM in Adam. See also Section~\ref{sec:adam>sgd}.} 
 	\label{fig:update_value}
\end{figure*}

\subsection{Preliminaries}
Binary neural network optimization is challenging because weights and activations of BNNs are discrete values in \{-1, +1\}. In particular, in the forward pass, real-valued weights and activations are binarized with the sign function. 
\begin{align}
a_b = {\rm Sign}(a_r) = & \left\{
             \begin{array}{lr}
             - 1 & {\rm if} \ \ a_r <0 \\
             + 1 & {\rm otherwise}
             \end{array}
\right. \\
\hspace{-0.7em}w_b = \frac{||W_r||_{l1}}{n} {\rm Sign}(w_r) & = \left\{
             \begin{array}{lr}
             - \frac{||W_r||_{l1}}{n} & {\rm if} \ w_r <0 \\ 
             \\
             + \frac{||W_r||_{l1}}{n} & {\rm otherwise}
             \end{array} 
\right. 
\end{align}
Note that, the real-valued activations $a_r$ are the outputs of the previous layers, generated by the binary or real-valued convolution operations. The real-valued weights $w_r$ are stored as {\em latent} weights to accumulate the small gradients. {\em Latent} refers to that the weights are not used in the forward pass computation. Instead, the sign of real-valued latent weights multiplying the channel-wise absolute mean ($\frac{1}{n} ||W_r||_{l1}$) is used for updating binary weights~\cite{rastegari2016xnor}.

In the backward pass, due to the non-differentiable characteristic of the sign function, the derivative of $clip(-1, a_r, 1)$ function is always adopted as the approximation to the derivative of the sign function~\cite{rastegari2016xnor}. 
It is noteworthy that, because the {\em sign} is a function with bounded range, the approximation to the derivative of the sign function will encounter a zero (or vanishing) gradient problem when the activation exceeds the effective gradient range ($[-1,1]$), which leads to the optimization difficulties that will be discussed in Section~\ref{sec:saturation}.

\subsection{Observations}

\subsubsection{Activation Saturation on Gradients} 
\label{sec:saturation}

{\em Activation saturation} is the phenomenon that the absolute value of activations exceeds one and the corresponding gradients are suppressed to be zero, according to the definition of approximation to the derivative of the sign function~\cite{ding2019regularizing}. From our observation, activation saturation exists in every layer of a binary network and it will critically affect the magnitude of gradients in different channels. In Figure~\ref{fig:activation_saturation}, we visualize the activation distributions of the first binary convolution layer. We can observe that many activations exceed the bounds of -1 and +1, making the gradient passing those nodes become zero-valued. According to the Chain Rule~\cite{ambrosio1990general}, the gradients are extremely vulnerable to the activation saturation in latter layers and thus will vibrate tempestuously in their corresponding magnitudes. 

\begin{figure*}[t]
	\centering
	    \includegraphics[width=0.98\textwidth]{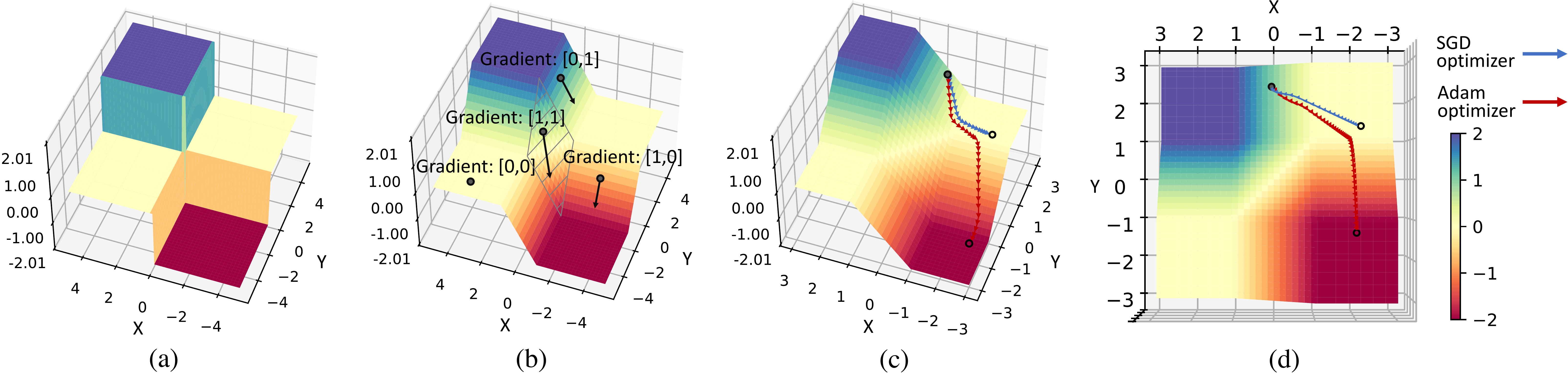}
	\vspace{-1em}
	\caption{The loss landscape visualization of a network constructed with the summation of two binary nodes. (a) the loss surface of the binary network in the forward pass, binarization functions $sign(x)$ discretized the landscape, (b) the loss surface for actual optimization after using the derivative of $clip(-1,x,1)$ in approximating the derivative of $sign(x)$, (c) the comparison between using SGD optimizer and Adam optimizer in conquering the zero gradient local minima, (d) the top view of the actual optimization trajectory.} 
	\label{fig:2D_loss}
\vspace{0.5em}
\end{figure*}

\subsubsection{Fairness in Weight Training}
\label{sec:fairness}

{\em Unfair training} is the phenomenon that the weights in some channels are not optimized to learn meaningful representations. Given different batches of images, the activation saturation usually occurs on different activation channels. In these channels, the gradient will always stay small in our observation, which causes unfair training.
Note that the weights refer to the real-valued latent weights in the binary neural network. The magnitude of these real-valued weights are regarded as \textit{`inertial'}~\cite{helwegen2019latent}, indicating how likely the corresponding binary weights are going to change their signs.

To measure the effect of unfair training, we calculate the Channel-wise Absolute Mean (CAM) to capture the average magnitude of real-valued weights within a kernel, which is represented as \textit{red} hyphens in Figure~\ref{fig:weight_distribution} and Figure~\ref{fig:update_value}. 
The definition of CAM is as follows:
\begin{align}
    \textrm{CAM} = \frac{1}{{N_{in}} \cdot k \cdot k} \sum_{c=1}^{N_{in}} \sum_{i=1}^k \sum_{j=1}^k | w_{\{c,i,j\}}| 
\end{align}
where $N_{in}$ is the number of input channels, $w$ is the weights in BNNs, $c$ is the channel index, $i, j$ are the element position in $c$-channel, and $k$ is the kernel size. We can see that when using SGD, the CAM of latent weights in a binary network are small in their values (Figure~\ref{fig:weight_distribution} (b)) compared with their real-valued counterparts (Figure~\ref{fig:weight_distribution} (a)) and is also higher in variance, which reflects the unbalanced weight training inside the SGD optimized binary network.

To measure the uniformness of the trained latent real-valued weight magnitude, we propose the Standard Deviation of the Absolute Mean (SDAM) of the real-valued weight magnitude on each output channel. The statistics of SDAM for SGD and Adam are shown in Figure~\ref{fig:weight_distribution}. It is evident that the SDAM of Adam is lower than that of SGD, revealing higher fairness and stability in the Adam training than SGD.

\subsubsection{Why is Adam Better than SGD?}
\label{sec:adam>sgd}

\label{sec:2D_loss}
For better illustration, we plot a two-dimensional loss surface of a network with two nodes where each node contains a sign function binarizing its input. As shown in Figure~\ref{fig:2D_loss} (a), the sign functions result in a discretized loss landscape with zero gradients at almost all input intervals, making the landscape infeasible to be optimized with gradient descent. 

In literature, the derivative of $clip(-1, a_r, 1)$ function is always adopted as the approximation to the derivative of the sign function. Thus, the actual landscape where gradients are computed is constructed with clip nodes. In Figure~\ref{fig:2D_loss} (b), the approximated gradients of binary activations retain their values in both direction only when both inputs land in the interval of [-1, 1], denoted as the slashed area in Figure~\ref{fig:2D_loss} (b). Outside this region, the gradient vector either has value in only one direction or contains zero value in both directions, which is the so-called flattened region.

During the actual BNN optimization, the activation value depends on the input images and will vary from batch to batch, which is likely to exceed [-1, 1]. This activation saturation effect in turn results in the gradient vanishing problem. For illustration, on this 2D-loss surface, we denote the starting point of optimization in grey circles. Started with the same sequence of gradients, the SGD optimizer computes the update value with the first momentum by definition:
$ v_t = \gamma v_{t-1} + g_t, $ 
where $g_t$ denotes the gradient and $v_t$ denotes the first momentum for weight update. While the update value in Adam is defined as:
$u_t = \frac{\hat{v}_t}{\sqrt{\hat{m}_t}+\epsilon},$ 
$\hat{v}_t$ and $\hat{m}_t$ denote exponential moving averages of the gradient and the squared gradient, respectively. At the flattened region, with $\hat{m}_t$ tracing the variance of gradients, the update value $u_t$ is normalized to overcome the difference in the gradient value. 
In contrast to SGD that only accumulates the first momentum, the adaptive optimizer, Adam, naturally leverages the accumulation in the second momentum to amplify the learning rate regarding the gradients with small historical values. As shown in Figure~\ref{fig:2D_loss} (c) and (d), Adam contains higher proportion in update value of $x$ direction compared to SGD when the gradient in $x$ direction vanishes. In our experiments, we found this property crucial for optimizing BNNs with more rugged surfaces and local flatten regions due to binarization. 
Figure~\ref{fig:update_value} also shows the update values of each iteration with CAM form in training an actual BNN. It confirms that with Adam training, the update values are usually larger than a threshold but with SGD, the values are very close to zero. As a result, ``dead'' weights from saturation are easier to be re-activated by Adam than SGD.

\subsubsection{Physical Meaning of Real-Valued Weight}
\label{sec:latent_weight}
The superiority of Adam for BNNs is also fortified in the final accuracy. As shown in Figure~\ref{fig:final_weight_distribution} (a), Adam achieves 61.49\% top-1 accuracy, comparing to 58.98\% of SGD in Figure~\ref{fig:final_weight_distribution} (b) with a consistent setting imposed on both experiments in terms of hyper-parameters and network structures.
Furthermore, we investigate the weight distribution in Figure~\ref{fig:final_weight_distribution} of final models and obtain some interesting discoveries. We find that the real-valued latent weights of better-performing models usually emerge to three peaks, one is around zero and the other two are beyond -1 and 1. For those poorly optimized models with SGD, the distributions of real-valued weights only contain one peak centering around zero. The physical significance of real-valued weights indicates the degree of how easy or difficult the corresponding binary weights can switch their signs (-1 or +1) to the opposite direction. If the real-valued weights are close to the central boundary ($0$), it will be simple for them to fall or bias to -1 or +1 through a few steps of gradient updating, making the whole network unstable. 
Thus, it is not far-fetched that real-valued weights can be regarded as the {\em confidence} of a binary value to be -1 or +1, as also being mentioned in~\cite{helwegen2019latent}. From this perspective, the weights learned by Adam are definitely more confident than those learned by SGD, which consistently verifies the conclusion that Adam is a better optimizer to use for binary neural networks.

\subsection{Metrics for Understanding BNN Optimization}
\label{sec:metrics}
Given the superiority of Adam over SGD, we take this finding further and investigate the training strategy for BNNs. Based on the intriguing fact that the BNN optimization relies on real-value weights for gradient accumulation and their signs for loss computation, BNN optimization is intractable compared to real-valued networks. Thus for better revealing the mechanism of the perplexing BNN training, we propose two metrics to depict the training process and further find that the weight decay added on the real-valued latent weight plays a non-negligible role in controlling the binary weights evolving.

\subsubsection{Weight Decay in BNN Optimization}

In a real-valued neural network, weight decay is usually used to regularize the real-valued weights from growing too large, which prevents over-fitting and helps to improve the generalization ability~\cite{krogh1992simple}.

However, for a binary neural network, the effect of weight decay is less straightforward. As the absolute values of weights in BNNs are restricted to -1 and +1, the weight decay is no longer effective to prevent the binary weights from being extremely large.
Moreover, in a binary neural network, the weight decay is applied to the real-valued latent weights. Recall that in Section~\ref{sec:latent_weight}, the magnitude of real-valued weights in BNNs can be viewed as the {\em confidence} of corresponding binary weights to their current values. Adding weight decay on these real-valued weights is actually attempting to decay the {\em confidence} score.

From this perspective, the weight decay will lead to a dilemma in binary network optimization between the stability and the dependency of weight initialization.
With high weight decay, the magnitude of the \textit{latent} weights is regularized to be small, making the corresponding binary weights ``less confident'' in their signs, and further prone to switch their signs frequently, i.e., reducing the stability in optimization. With smaller or even zero weight decay, the {\em latent} weights tend to move towards -1 and +1, the corresponding binary weights will be more stable to stay in the current status. However, this is a trade-off since larger gradients are required to promote the weights to switch their signs in order to overcome the ``dead'' parameters issue. That is to say, with small or zero weight decay, the performance of a network will be influenced by initialization critically.

\begin{figure}[t]
	\centering
	{
	\begin{subfigure}[b]{0.236\textwidth}
	    \centering
	    \includegraphics[height=0.63\textwidth]{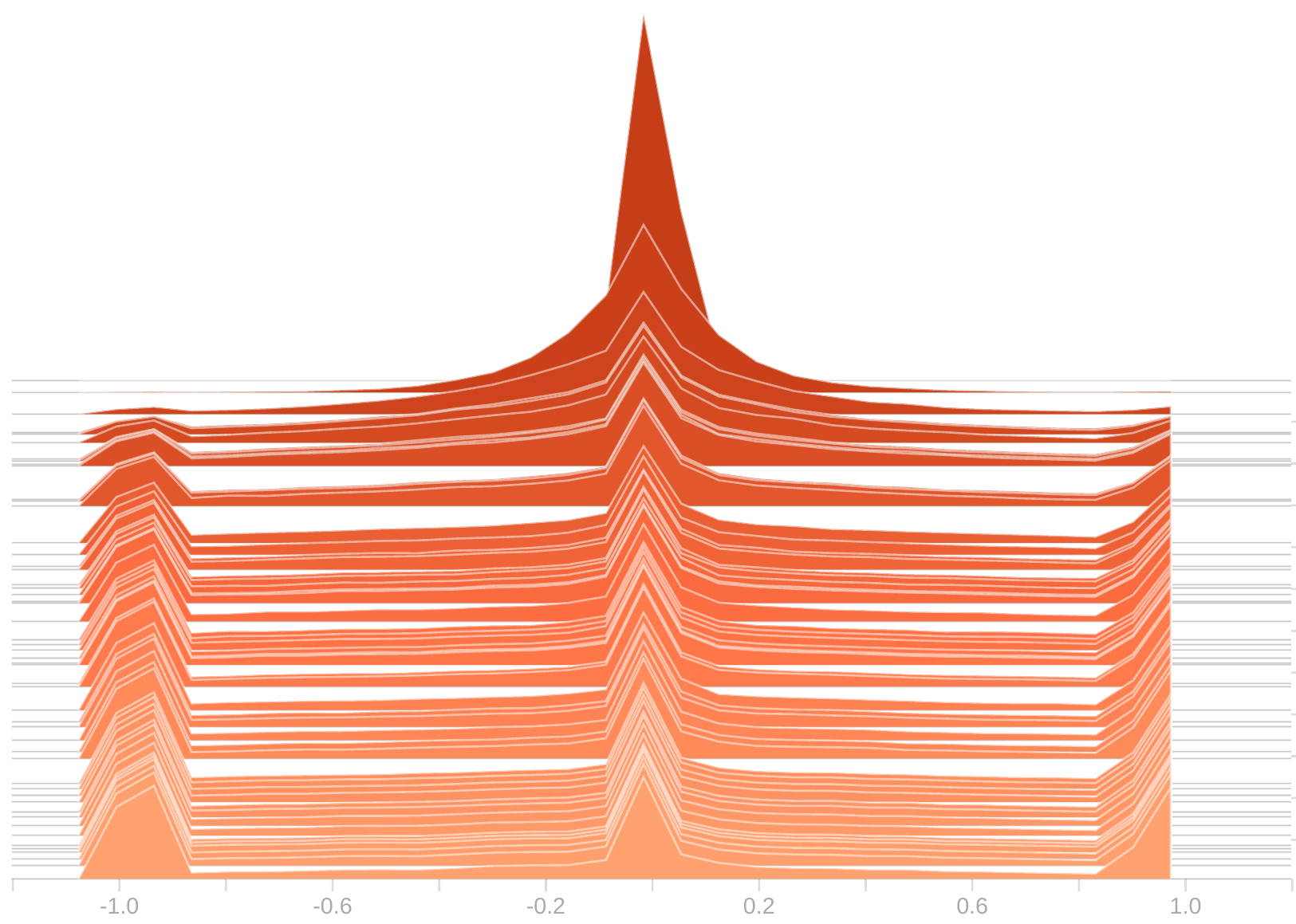}
	    \caption{\centering
        Adam with one step training Top1-Acc: 61.49\%} 
	\end{subfigure}
	\begin{subfigure}[b]{0.236\textwidth}
	    \centering
	    \includegraphics[height=0.63\textwidth]{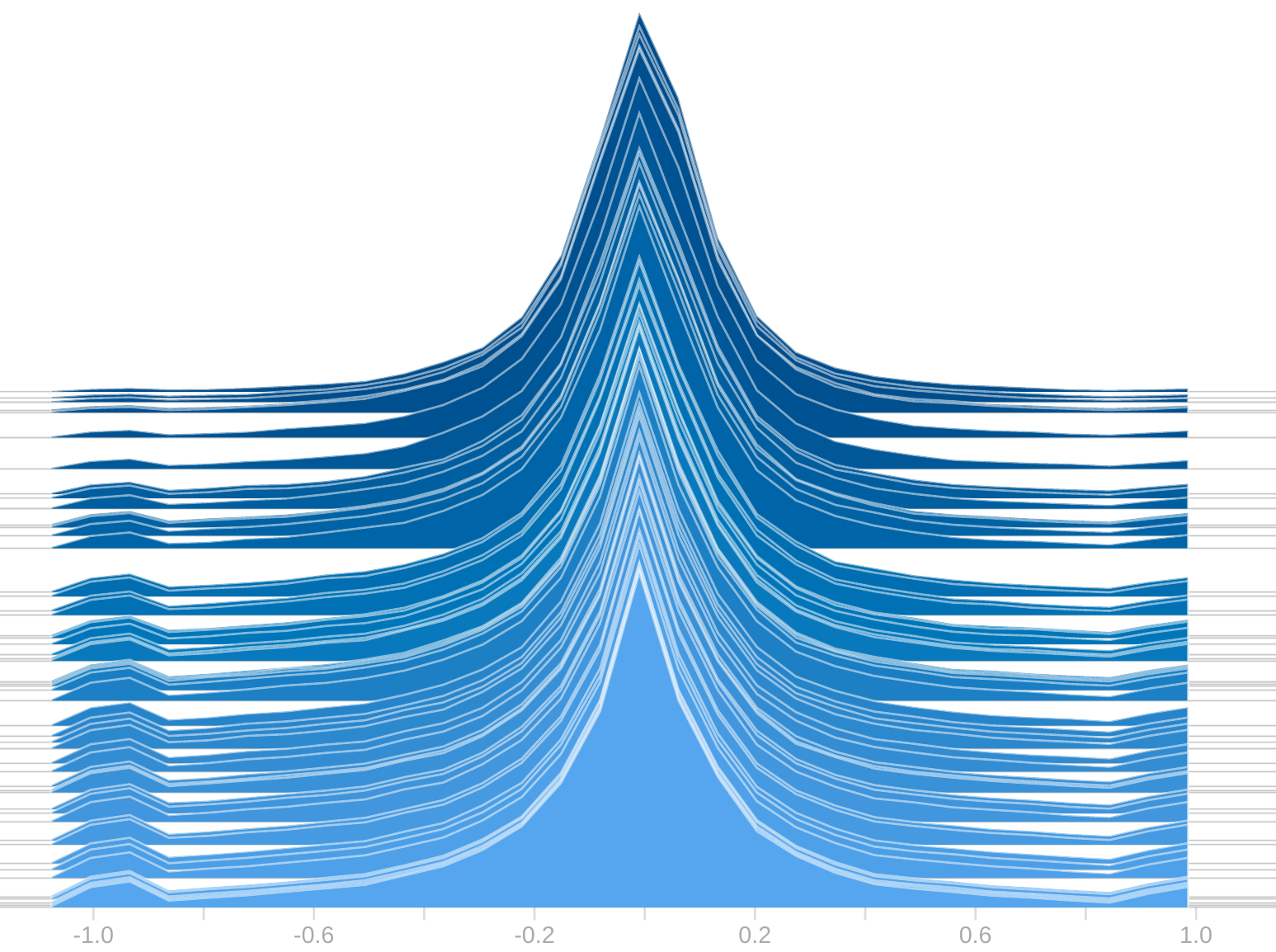}
	    \caption{\centering
        SGD with one step training Top1-Acc: 58.98\%} 
	\end{subfigure}
		\begin{subfigure}[b]{0.236\textwidth}
	    \centering
        \includegraphics[height=0.63\textwidth]{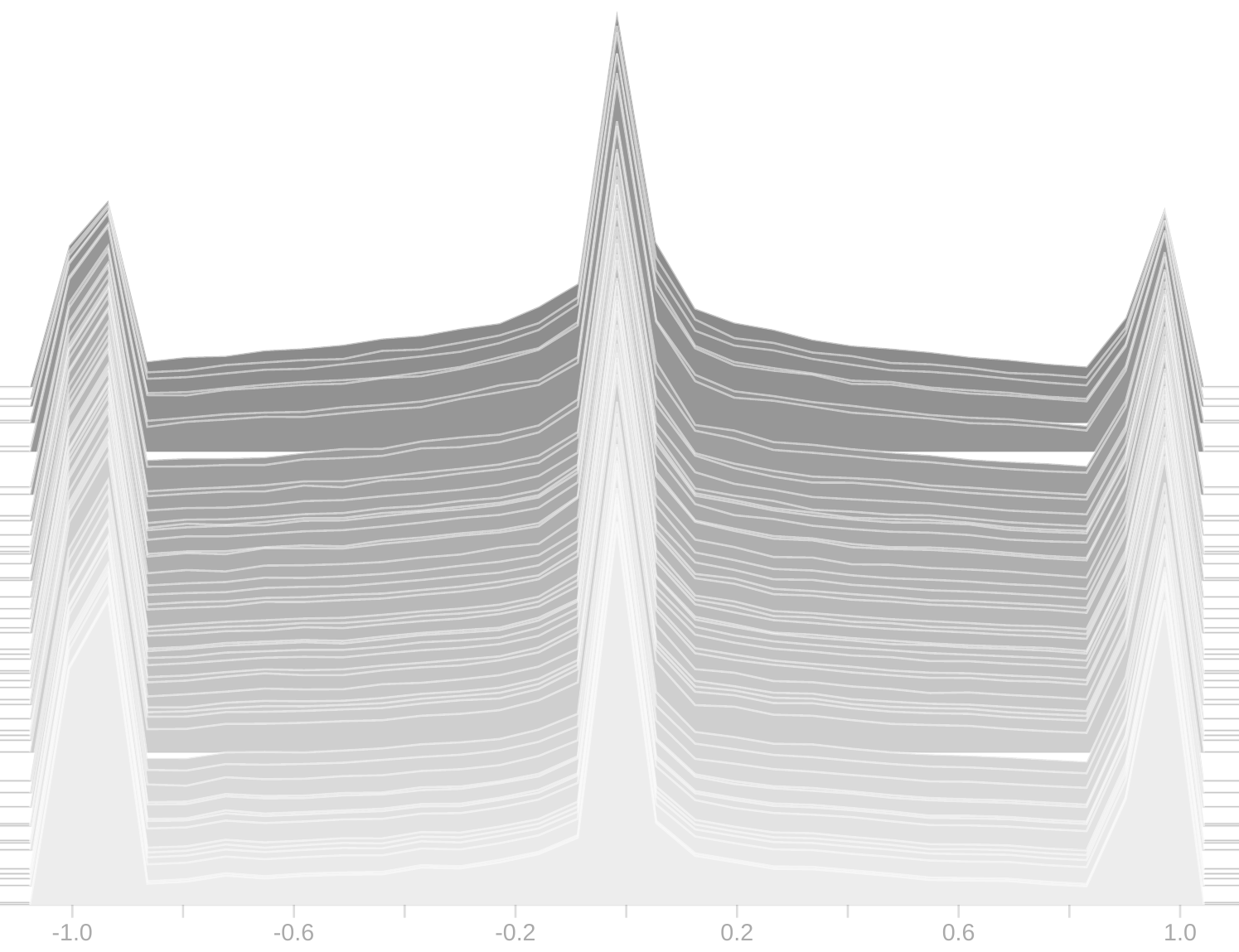}
        \caption{\centering
        Adam with two step training Top1-Acc: 63.23\%}
	\end{subfigure}
		\begin{subfigure}[b]{0.236\textwidth}
	    \centering
        \includegraphics[height=0.63\textwidth]{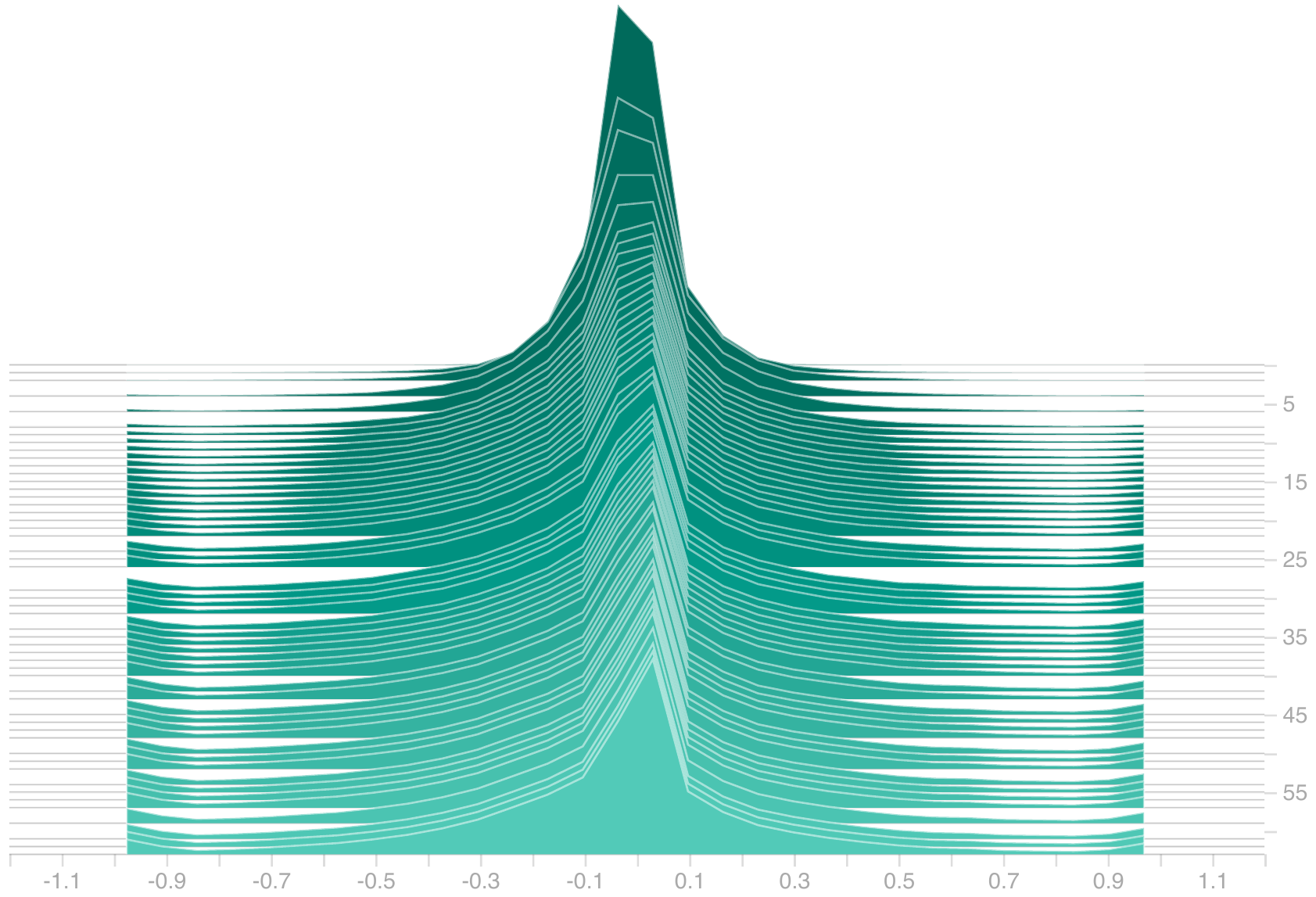}
        \caption{\centering
        SGD with two step training Top1-Acc: 61.25\%}
	\end{subfigure}\\
	}
	\vspace{-0.8em}    
	\caption{The final weight distribution. We found that Adam has more latent real-valued weights with larger absolute values compared to SGD. Since the real-valued weights can be viewed as the {\em confidence score} of the corresponding binary weights in their current sign, Adam-optimized binary networks are more confident in values than that of SGD, and the final accuracy is also higher.} 
	\label{fig:final_weight_distribution}
\end{figure}

\subsubsection{Quantification Metrics}
To quantify these two effects (network stability and initialization dependency), we introduce two metrics: the flip-flop (FF) ratio for measuring the optimization stability, and the correlation-to-initialization (C2I) ratio for measuring the dependency on initialization.
The FF ratio is defined as:
\begin{align}
    & {\bf I_{FF}} = \frac{|\textrm{Sign}(w_{t+1})-\textrm{Sign}(w_{t})|_\textit{abs}}{2},\\
    & {\bf FF_{ratio}} = \frac{\sum_{l=1}^L\sum_{w \in W_l} {\bf I_{FF}}}{N_{\textrm{total}}},
\end{align}
where ${\bf I_{FF}}$ is the indicator of whether a weight changes its sign after the updating at iteration $t$. $N_{total}$ is the total number of weights in a network with $L$ convolutional layers. ${\bf FF_{ratio}}$ denotes the ratio of flip-flops, \textit{i.e.,} percentage of weights that change their signs.

Then we define C2I ratio as:
\begin{align}
    & {\bf I_{C2I}} = \frac{|\textrm{Sign}(w_{\textrm{final}})-\textrm{Sign}(w_{\textrm{init}})|_\textit{abs}}{2}, \\
    & {\bf C2I_{ratio}} = 1 - \frac{1}{2} \frac{\sum_{l=1}^L\sum_{w \in W_l} {\bf I_{C2I}}}{N_{\textrm{total}}},
\end{align}
where ${\bf I_{C2I}}$ is the indicator of whether a weight has different sign to its initial sign and ${\bf C2I_{ratio}}$ denotes the correlation between the signs of final weights and the initial values.

Here we study the FF ratio and C2I ratio for different weight decay values. 
From Table~\ref{table:FF_and_C2I}, it is easy to find that the FF ratio is in negative correlation with C2I ratio. With the increase of weight decay, the FF ratio increases exponentially while the C2I ratio decreases linearly. This indicates that some flip-flops do not contribute to the final weights, but just harm the training stability. 

In this experiment, we found using the weight decay of 5e-6 produces the highest accuracy. Further, we discover that a particular two-step training scheme~\cite{martinez2020training, liu2020reactnet} can disentangle the negative correlation between FF ratio and C2I ratio.

\setlength{\tabcolsep}{2pt}
\begin{table}[t]
\begin{center}
\caption{The FF ratio, C2I ratio and Top-1 accuracy by Adam optimization with different weight decay. Note that the FF ratios in this table are averaged over the total training iterations.}
\vspace{-1em}
\label{table:FF_and_C2I}
\resizebox{0.49\textwidth}{!}{
\hspace{-1em}
\begin{tabular}{cccccc}
\noalign{\smallskip}
\hline\noalign{\smallskip}
\hline
& Weight decay & FF ratio & C2I ratio & Top1-acc\\
\hline
\multirowcell{4}{One-step}
& 1e-5 & 2.33$\times$1e-3 & 0.4810 & 61.73\\
& 5e-6 & 1.62$\times$1e-3 & 0.4960 & \textbf{61.89}\\
& 0 & 2.86 $\times$1e-4 & 0.5243 & 61.49 \\
& -1e-4 & 1.07 $\times$1e-7 & 0.9740 & 26.21 \\
\hline
\multirowcell{2}{Two-step} & {\em Step1:} 1e-5 \ \ {\em Step2:} 0 & 4.89$\times$1e-4 & 0.6315 & 62.63 \\
& {\em Step1:} 5e-6 \ \ {\em Step2:} 0 & 4.50$\times$1e-4 & 0.6636 & \textbf{63.23}\\
\hline
\hline
\end{tabular}
}
\end{center}
\end{table}
\subsubsection{Practical Training Suggestion}
Intrinsically, the dilemma of whether adding weight decay on real-valued latent weights originates from the fact that the binary weights are discrete in value. For real-valued latent weights around zero, a slight change in value could result in a significant change in the corresponding binary weights, thus making it fairly tricky to encourage real-valued latent weights to gather around zero. 

Interestingly, we found that a good weight decay scheme for the recent two-step training algorithm~\cite{martinez2020training,liu2020reactnet} can disentangle this dilemma. In {\em Step1}, only activations are binarized and the real-valued weights with weight decay are used to accumulate small update values. Since real-valued networks have no worries about the FF ratio, we can simply add weight decay to harvest the benefit of low initialization dependency. Then, in {\em Step2}, we initialize latent real weights in the binary networks with weights from {\em Step1}, and enforce a weight decay of 0 on them. With this operation, we can reduce the FF ratio to improve stability and utilize the good initialization from {\em Step1} (similar to pre-training) rather than the random parameters. In this stage, a high C2I ratio will not harm the optimization. From this perspective, we found that 5e-6 as weight decay performs best for balancing the weight magnitude for a good initialization in {\em Step2}.

As shown in Figure~\ref{fig:final_weight_distribution} (c), more real-valued weights in two-step training tend to gather around -1 and +1, indicating that this strategy is more confident than one-step. By simply eliminating the undesirable weight decay value just by looking at the FF ratio in the early epochs we can find a good weight decay with fewer trials and errors. We will see in Section~\ref{sec:experiments} that our training strategy outperforms the state-of-the-art ReActNet by 1.1\% with identical architectures. 

\setlength{\tabcolsep}{1pt}
\begin{table}[t]
\begin{center}
\caption{Comparison with state-of-the-art methods that binarize both weights and activations.}
\label{table:SOTA}
\resizebox{0.45\textwidth}{!}{
\begin{tabular}{cccccccc}
\noalign{\smallskip}
\hline
\hline
\multirowcell{2}{Networks} & Top1 & Top5 \\
& Acc \% & Acc \% \\
\hline
BNNs~\cite{courbariaux2016binarized}& 42.2 & 67.1\\
ABC-Net~\cite{lin2017abcnet}& 42.7 & 67.6\\
DoReFa-Net~\cite{zhou2016dorefa}& 43.6 & -\\
XNOR-ResNet-18~\cite{rastegari2016xnor}& 51.2 & 69.3\\
Bi-RealNet-18~\cite{liu2018bi}& 56.4 & 79.5 \\
CI-BCNN-18~\cite{wang2019ci-bcnn}& 59.9 & 84.2 \\
MoBiNet \cite{Hai_2020_WACV} & 54.4 & 77.5 \\
BinarizeMobileNet~\cite{phan2020binarizing} & 51.1 & 74.2 \\
PCNN ~\cite{gu2019projection}& 57.3 & 80.0 \\
StrongBaseline~\cite{martinez2020training}& 60.9 & 83.0 \\
Real-to-Binary Net~\cite{martinez2020training} & 65.4 & 86.2\\
MeliusNet29~\cite{bethge2020meliusnet} & 65.8 & --\\
ReActNet ResNet-based~\cite{liu2020reactnet} & 65.5 & 86.1 \\
ReActNet-A~\cite{liu2020reactnet} & 69.4 & 88.6 \\
\hline
StrongBaseline + Our training strategy & 63.2 & 84.0 \\
ReActNet-A + Our training strategy & \textbf{70.5} & \textbf{89.1}\\
\hline
\hline
\end{tabular}
}
\vspace{-0.28in}
\end{center}
\end{table}

\setlength{\tabcolsep}{1pt}
\begin{table}[t]
\begin{center}
\caption{Comparison of computational cost between the state-of-the-art methods and our method.}
\label{table:FLOPs}
\resizebox{0.48\textwidth}{!}{
\begin{tabular}{cccccccc}
\noalign{\smallskip}
\hline
\hline
\multirowcell{2}{Networks}& \multirowcell{2}{BOPs \\ $\times10^9$} & \multirowcell{2}{FLOPs \\ $\times10^8$} & \multirowcell{2}{OPs\\$\times10^8$}\\
\\
\hline
XNOR-ResNet-18~\cite{rastegari2016xnor}& $1.70$ & $1.41$ & $1.67$\\
Bi-RealNet-18~\cite{liu2018bi}& $1.68$ & $1.39$ & $1.63$\\
CI-BCNN-18~\cite{wang2019ci-bcnn}& -- & -- & $1.63$\\
MeliusNet29~\cite{bethge2020meliusnet} & $5.47$ & $1.29$ & $2.14$ \\
StrongBaseline~\cite{martinez2020training} & $1.68$ & $1.54$ & $1.63$\\
Real-to-Binary~\cite{martinez2020training} & $1.68$ & $1.56$ & $1.83$ \\
ReActNet-A~\cite{liu2020reactnet} & $4.82$ & $0.12$ & $0.87$ \\
\hline
StrongBaseline + Our training strategy & $1.68$ & $1.54$ & $1.80$\\
ReActNet-A + Our training strategy & $4.82$ & $0.12$ & $0.87$\\
\hline
\hline
\end{tabular}
}
\vspace{-0.18in}
\end{center}
\end{table} 

\begin{table}[t]
\begin{center}
\caption{Comparison of different binarization orders in two-step training on the StrongBaseline~\cite{martinez2020training} structure.}
\label{table:order}
\resizebox{0.43\textwidth}{!}{
\begin{tabular}{cccccc}
\noalign{\smallskip}
\hline
\hline
 & Top1 Acc & \ \ Top5 Acc \ \ \\
\hline
first binarize weight & \multirowcell{2}{60.17} & \multirowcell{2}{82.05} \\
then binarize activation (\textit{BWBA}) \\
\hline
first binarize activation & \multirowcell{2}{\textbf{63.23}} & \multirowcell{2}{\textbf{84.02}} \\
then binarize weight (\textit{BABW}) \\
\hline
\hline
\end{tabular}
}
\vspace{-0.28in}
\end{center}
\end{table}

\begin{table}[t]
\begin{center}
\caption{Comparison between Adam and other adaptive methods.}
\label{table:adam_vs_other_adaptive}
\resizebox{0.4\textwidth}{!}{
\begin{tabular}{cccccccc}
\noalign{\smallskip}
\hline
\hline
& \multirowcell{2}{Adam} & \multirowcell{2}{RMS-\\prop} & \multirowcell{2}{Ada-\\Grad} & \multirowcell{2}{Ada-\\Delta} & \multirowcell{2}{AMS-\\Grad} & \multirowcell{2}{Ada-\\Bound} \\
\\
\hline
Top1-acc \ & \textbf{61.49} \ & 57.90 \ & 50.74 \ & 56.90 \ & 60.71 \ & 58.13 \ \\
Top5-acc \ & \textbf{83.09} \ & 79.93 \ & 74.62 \ & 79.47 \ & 82.44 \ & 80.58 \ \\
\hline
\hline
\end{tabular}
}
\end{center}
\vspace{-0.18in}
\end{table}
\setlength{\tabcolsep}{1.4pt}

\section{Experiments}
\label{sec:experiments}

\subsection{Dataset and Implementation Details}
\label{sec:dataset}
All the analytical experiments are conducted on the ImageNet 2012 classification dataset~\cite{imagenet}. 
We train the network for 600K iterations with batch size set to 512. The initial learning rate is set to 0.1 for SGD and 0.0025 for Adam, with linear learning rate decay. We also adopt the same data augmentation in~\cite{martinez2020training} and the same knowledge distillation scheme as~\cite{liu2020reactnet} for training ReActNet structures.
For a fair comparison of optimization effects, we use the same network structures as StrongBaseline in ~\cite{martinez2020training} for all the illustrative experiments and compared our training strategy on two state-of-the-art network structures including StrongBaseline, and ReActNet~\cite{liu2020reactnet}.

\vspace{-0.06in}
\subsection{Comparison with State-of-the-Arts}
\label{sec:SOTA}
\vspace{-0.03in}
Our training strategies bring constant improvements to both structures.
As shown in Table~\ref{table:SOTA}. With the same network architecture, we achieve 2.3\% higher accuracy than the StrongBaseline~\cite{martinez2020training}. When applying our training strategy to the state-of-the-art ReActNet~\cite{liu2020reactnet}, it further brings 1.1\% enhancement and achieves 70.5\% top-1 accuracy, surpassing all previous BNN models.

Our training strategy will not increase the OPs as we use identical structures as the baselines: StrongBaseline~\cite{martinez2020training} and ReActNet~\cite{liu2020reactnet}. 
Table~\ref{table:FLOPs} shows the computational costs of the networks we utilized in experiments. StrongBaseline is a ResNet-18 based binary neural network, and it has similar OPs as Bi-RealNet-18~\cite{liu2018bi} and Real-to-Binary Network~\cite{martinez2020training}. ReActNet is a MobileNet-based BNN, and it contains small overall OPs than other binary networks.

\subsection{Ablation Study}
\label{sec:ablation_study}
\subsubsection{Comparison of Adam and SGD under Different Learning Rates}
\begin{figure}[t]
 	\vspace{-1em}
    \includegraphics[width=0.45\textwidth]{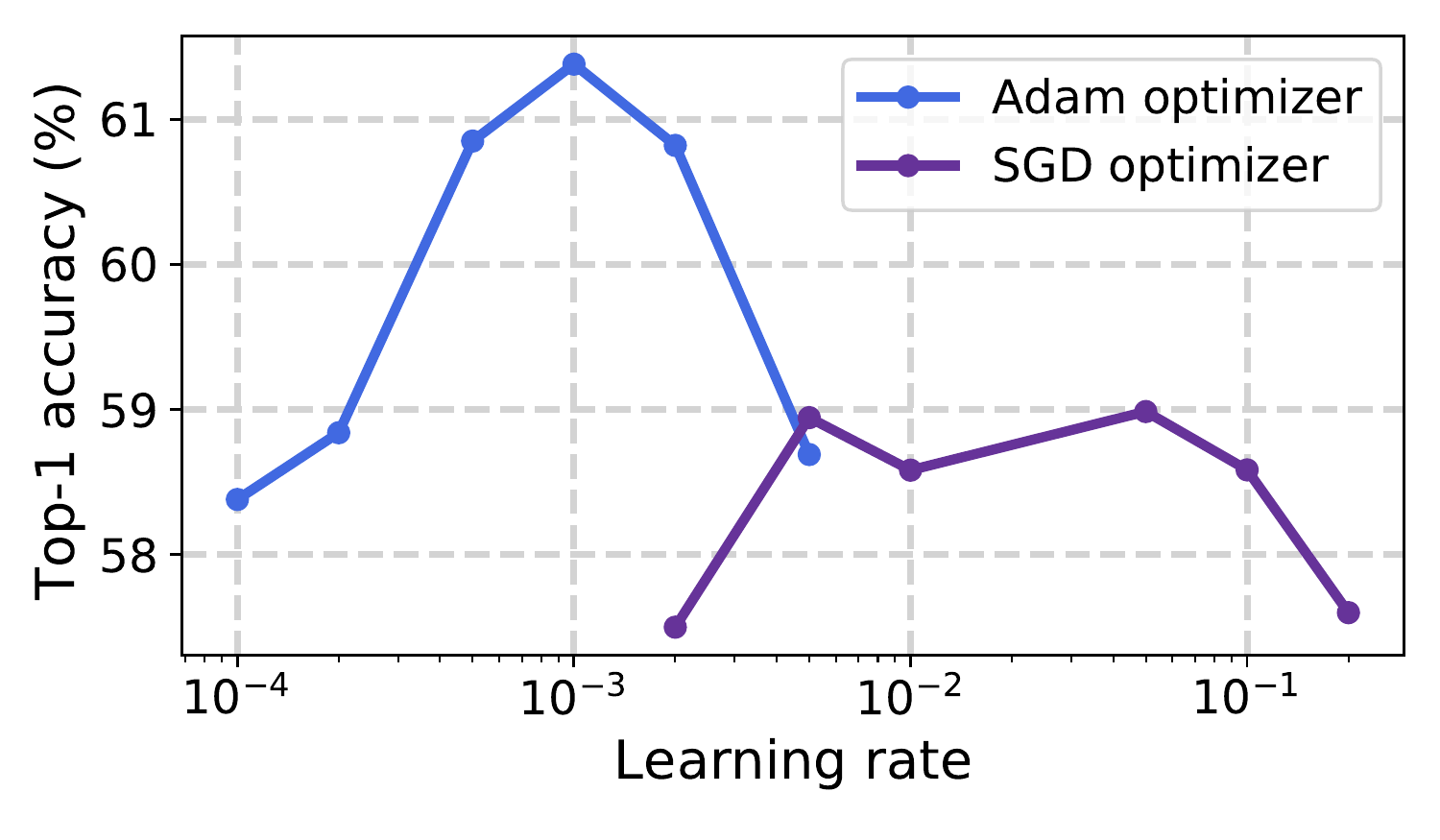}
 	\vspace{-1.7em}
	\caption{Accuracy {\em vs.} initial learning rate on Adam and SGD.}
	\label{fig:different_lr_curve}
\end{figure}

In Figure~\ref{fig:different_lr_curve}, we illustrate the Top-1 accuracy curves with different learning rates. To control variables, experiments are done with one-step training strategy on the ImageNet dataset with the StrongBaseline~\cite{martinez2020training} structure. In general, Adam can achieve higher accuracy across a variety of learning rate values and is also more robust than SGD. Besides, we observe that Adam enjoys small learning rates. The reason is that Adam adopts the adaptive method to update the gradients, which will amplify the actual learning rate values during training, so it requires a smaller initial learning rate to avoid update values being too large.

\subsubsection{Two-step Training}
To reassure the credibility of choosing the suggested two-step training algorithm, we make a controlled comparison between different training schemes. In Table~\ref{table:order}, our suggested order which first binarizes activations then weights (\textit{BABW}) obtained a 2.93\% better accuracy over the reversed order (\textit{BWBA}). In \textit{BWBA}, binary weights are adopted in both steps, which are confined to be discrete. So compared to the real-valued weights in \textit{Step1} of \textit{BABW}, it is harder for binary weights in \textit{Step1} of \textit{BWBA} to be well optimized for delivering a good initialization for \textit{Step2}. Thus \textit{BWBA} can not achieve the effect of breaking the negative correlation between FF ratio and C2I ratio.

\subsubsection{Comparison with Other Adaptive Methods}
\label{sec:other_adaptive}

In this experiment, the initial learning rates for different optimizers are set to the PyTorch~\cite{paszke2019pytorch} default values (0.001 for Adam, 0.01 for RMSprop, 0.01 for AdaGrad, 1.0 for AdaDelta, and 0.001 for AMSGrad). For Adabound, we adopt the default learning rate schedule in~\cite{luo2019adabound} by setting the initial learning rate to 0.001 and transiting to 0.1. The weight decay is set to 0. For fair comparison, these experiments are carried out with one-step training on the ImageNet dataset with the StrongBaseline~\cite{martinez2020training} structure. 

In Table~\ref{table:adam_vs_other_adaptive}, Adam~\cite{kingma2014adam} achieves similar accuracy with its variant AMSGrad~\cite{reddi2019amsgrad}, and better results than other adaptive methods. RMSprop~\cite{rmsprop} and Adadelta~\cite{zeiler2012adadelta} are adaptive methods without using the first momentum of the gradients. In binary neural networks, since the gradients with respect to the discrete weights are noisy, the first momentum is also crucial for averaging out the noise and improve the accuracy. AdaGrad~\cite{duchi2011adagrad} is known that its accumulation of the squared gradients in the denominator will keep growing during training, causing the learning rate to shrink and eventually become infinitesimally small, and preventing the algorithm to acquire additional knowledge. Thus the performance of AdaGrad is modest. As a variant of Adam, AMSGrad uses ``long-term memory'' of the past gradients to avoid extreme adaptive learning rate, which achieves comparable accuracy as Adam on a binary classification network, while AdaBound~\cite{luo2019adabound} is proposed to smoothly transit from Adam to SGD in order to harvest the good generalization ability of SGD at the end of training. However, in binary neural network optimization, SGD does not show its superiority in improving the generalization as in real-valued networks. But instead, in BNNs optimization, transiting to SGD leads to unstableness in training and failure in dealing with extremely small gradients, which leads to a worse accuracy. 

\section{Conclusion and Future Work}

Many state-of-the-art BNNs are optimized with Adam, but the essential relations between BNNs and Adam are still not well-understood. In this work, we made fair comparisons between Adam and SGD for optimizing BNNs. We explain how Adam helps to re-activate those ``dead'' weights for better generalization. All our explanations are reflected in the visualization results. Furthermore, we elucidate why weight decay and initialization are critical for Adam to train BNNs and how to set their values. As we have shown, with the appropriate scheme of two-step training, our method achieved a competitive result of 70.5\% on ImageNet. We hope these findings and understandings can inspire more studies in BNN optimization. Our future work will focus on designing new optimizers specifically for binary networks.

\bibliography{example_paper}
\bibliographystyle{icml2021}
\end{document}